\documentclass[10pt, a4paper]{article}
\usepackage{lrec}
\usepackage{graphicx}
\usepackage{tabularx}
\usepackage{soul}
\usepackage[latin1]{inputenc}
\usepackage{hyperref}
\usepackage{xstring}

\title{Building a Knowledge Graph from Natural Language Definitions \\ for Interpretable Text Entailment Recognition}

\name{Vivian S. Silva\textsuperscript{1}, Andr\'{e} Freitas\textsuperscript{2}, Siegfried Handschuh\textsuperscript{1}}

\address{\textsuperscript{1}Department of Computer Science and Mathematics, University of Passau,	Innstra\ss{}e 43, 94032, Passau, Germany \\	\textsuperscript{2}School of Computer Science, University of Manchester, Kilburn Building,  Oxford Road, M13 9PL, UK \\ 	vivian.santossilva@uni-passau.de, andre.freitas@manchester.ac.uk, siegfried.handschuh@uni-passau.de
\vspace{10pt}}

\abstract{
Natural language definitions of terms can serve as a rich source of knowledge, but structuring them into a comprehensible semantic model is essential to enable them to be used in semantic interpretation tasks. We propose a method and provide a set of tools for automatically building a graph world knowledge base from natural language definitions. Adopting a conceptual model composed of a set of semantic roles for dictionary definitions, we trained a classifier for automatically labeling definitions, preparing the data to be later converted to a graph representation. WordNetGraph, a knowledge graph built out of noun and verb WordNet definitions according to this methodology, was successfully used in an interpretable text entailment recognition approach which uses paths in this graph to provide clear justifications for entailment decisions. \\ \newline  }

\begin{document}

\maketitleabstract

\section{Introduction}\label{sec:intro}

Natural language lexical definitions of terms can be used as a source of knowledge in a number of semantic tasks, such as Question Answering, Information Extraction and Text Entailment. While formal, structured resources such as ontologies are still scarce and usually target a very specific domain, a large number of linguistic resources gathering dictionary definitions is available not only for particular domains, but also addressing wide-coverage commonsense knowledge.

However, in order to make the most of those resources, it is necessary to capture the semantic shape of natural language definitions and structure them in a way that favors both the information extraction process and the subsequent information retrieval, allowing the effective construction of semantic models from these data sources while keeping the resulting model easily searchable and interpretable. Furthermore, by using these models, systems can increase their own interpretability, benefiting from the structured data for performing traceable reasoning and generating explanations -- features which are becoming even more valuable given the growing importance of Explainable AI \cite{gunning2017explainable}.

In this work, we propose a method for automatically building commonsense knowledge bases out of natural language dictionary definitions, which is easily extensible to any domain where natural language glossaries are available. Building upon a conceptual model based on a set of semantic roles for definitions, we classify each segment in a definition according to its relation to the entity being defined, and convert the classified data into a knowledge graph where each node is a meaningful phrase which contains a piece of self-contained information about the definiendum. Following this methodology, we processed the whole noun and verb databases of WordNet \cite{fellbaum1998wordnet} and built the WordNetGraph, and then used this knowledge graph to recognize text entailments in an interpretable way, providing concise justifications for the entailment decisions.

\section{Related Work}\label{sec:relwork}

The construction of structured databases from dictionary definitions has been largely explored, and most approaches rely on syntactic parsers for the identification of patterns that point to relationships between words \cite{calzolari1991acquiring,vossen1991converting,vossen1992automatic,vossen1994untangling}. Among early efforts, it is remarkable the creation of LKB, a Lexical Knowledge Base \cite{copestake1991lkb} based on typed-feature structures that can be seen as a set of attributes for a given concept, such as ``origin'', ``color'', ``smell'', ``taste'' and ``temperature'' for the concept \textit{drink}, for example. The definitions from a machine-readable dictionary are parsed to extract the definiendum's genus and differentiae, and the values represented by the differentiae will fill in the feature structures for that genus. Since the features, that is, the relevant attributes for a given entity, must be defined in advance, only a restricted domain was considered in their approach.

Dolan et al. \shortcite{dolan1993automatically} also describe an automated strategy to build a structured lexical knowledge base but, instead of the entity-attributes structure, they use syntactic parsing to identify semantic relations such as \textit{is-a}, \textit{part-of}, etc., to build a directed graph. Recski \shortcite{recski2016building} also derives a graph representation from dictionary definitions, but in the adopted conceptual model there are only three types of edges, numbered from 0 to 2: the 0-edge represents unary predicates and the 1 and 2-edges connects binary predicates to their arguments. In common, most approaches work at the word-level, converting each single word in the definition into a different attribute or node. In the graph knowledge base scenario, this can increase the information retrieval complexity, given that it may be necessary to concatenate the contents of several nodes to obtain meaningful enough information about an entity.

The work proposed by \cite{bovi2015large} go beyond the word-level representation, being able to identify multi-word expressions. They perform a syntactic-semantic analysis of textual definitions for Open Information Extraction (OIE). Although they generate a syntactic-semantic graph representation of the definitions, the resulting graphs are used only as an intermediary resource for the final goal of extracting semantic relations between the entities present in the definition.

\section{Graph Conceptual Model}\label{sec:model}

To build the definition graph, we adopted the conceptual model proposed by Silva et al. \shortcite{silva2016categorization}. This model extends the genus-differentia definition pattern from Aristotle's classic theory of definition \cite{berg1982aristotle,lloyd1962genus,granger1984aristotle} by defining a set of entity-centered \textit{semantic roles} for lexical definitions. Differently from the commonly used event-centered semantic roles, which define the semantic relations holding among a predicate (the main verb in a clause) and its associated participants and properties \cite{marquez2008semantic}, definition's semantic roles express the part played by an expression in a definition, showing how it relates to the \textit{definiendum}, that is, the entity being defined.

\begin{figure*}[t]
	\begin{center}
		\includegraphics[width=6.3in, height=3.18in]{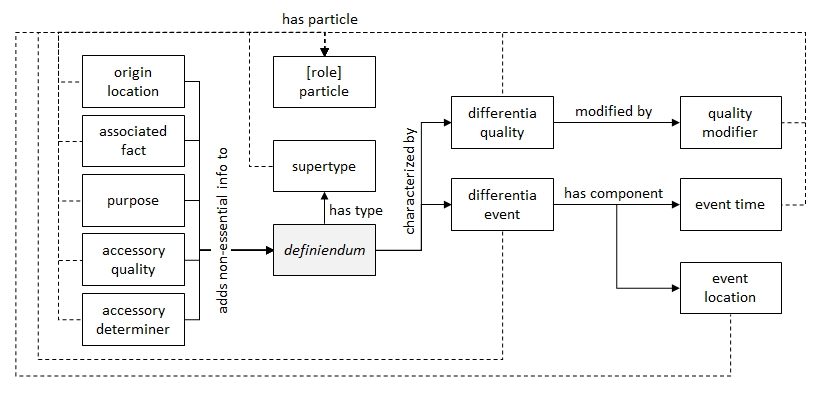} 
		\caption{Conceptual model for the semantic roles for lexical definitions. Relationships between \textit{[role] particle} and every other role in the model are expressed as dashed lines for readability.}
		\label{fig:sem_model}
	\end{center}
\end{figure*}

In this model, the \textit{genus} concept was replaced by the more general role \textit{supertype}, which can be not only the definiendum's immediate superclass but also an ancestor higher in the concepts hierarchy. The \textit{differentia} component was split into two roles: \textit{differentia quality} and \textit{differentia event}. These three roles can be seen as the representatives of an entity's essential properties, while other roles, such as \textit{associated fact}, \textit{purpose} or \textit{accessory quality}, for example, define non-essential properties. The conceptual model is depicted in Figure \ref{fig:sem_model}, and Table \ref{tab:roles} presents a summarized description for each of the roles defined in this model.

\begin{table}[h]
	\centering
	\begin{tabularx}{\linewidth}{l X}
		\textbf{Role}        & \textbf{Description}                                                                                                                                  \\ \hline
		Supertype            & the immediate or ancestral entity's superclass                                                                                                        \\
		Differentia quality  & a quality that distinguishes the entity from the others under the same supertype                                                                      \\
		Differentia event    & an event (action, state or process) in which the entity participates and that is mandatory to distinguish it from the others under the same supertype \\
		Event location       & the location of a differentia event                                                                                                                   \\
		Event time           & the time in which a differentia event happens                                                                                                         \\
		Origin location      & the entity's location of origin                                                                                                                       \\
		Quality modifier     & degree, frequency or manner modifiers that constrain a differentia quality                                                                            \\
		Purpose              & the main goal of the entity's existence or occurrence                                                                                                 \\
		Associated fact      & a fact whose occurrence is/was linked to the entity's existence or occurrence                                                                         \\
		Accessory determiner & a determiner expression that doesn't constrain the supertype-differentia scope                                                                        \\
		Accessory quality    & a quality that is not essential to characterize the entity                                                                                            \\
		{[}\textit{Role}{]} particle  & a particle, such as a phrasal verb complement, non-contiguous to the other role components                                                           
	\end{tabularx}
	\caption{Semantic roles for dictionary definitions}
	\label{tab:roles}
\end{table}

This set of semantic roles captures the semantic ``shape'' of natural language definitions and allows the extraction of structured representations from linguistic resources, enabling them to be used as knowledge sources in a wide range of semantic tasks.

\section{Construction Methodology}\label{sec:method}

\begin{figure*}[ht]
	\centering
	\includegraphics[width=6.5in, height=0.65in]{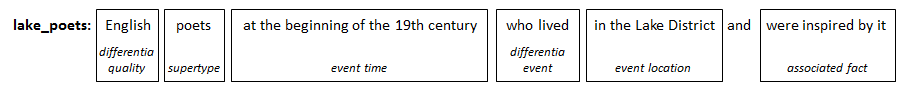}
	\caption{Example of role labeling for the definition of the ``lake\_poets'' synset.}
	\label{fig:classif_example}
\end{figure*}

Structuring natural language definitions as a graph allows us to select the portions of information regarding an entity's description that are relevant for a certain reasoning task. For example, consider the definition (from WordNet) for the concept ``lake poets'', which was classified according to the model described in Section \ref{sec:model}, illustrated in Figure \ref{fig:classif_example}. When retrieving data related to this concept, we could be interested only in origin- (\textit{lake poets are English poets}), time- (\textit{lake poets are poets at the beginning of the 19th century}) or space- (\textit{lake poets are poets who lived in the Lake District}) related information. When each of those roles is represented as a node in a graph we can focus only on the path containing the nodes of interest. Moreover, since the definition is split into segments rather than single words, each node contains a comprehensible amount of information, avoiding the need to visit several nodes to gather intelligible phrases.

To generate the \textit{WordNetGraph}\footnote{https://github.com/Lambda-3/WordnetGraph} -- a knowledge graph following the RDF data model -- from WordNet's noun and verb glosses, we adopted the following methodology for classifying and structuring the definitions:

\textbf{Synsets sample selection:} in order to use a supervised machine learning model to classify the data, we needed a initial set of annotated definitions. To build this set, we randomly selected 2,000 WordNet synsets, being 1,732 noun synsets and 268 verb synsets (the verb database size is around 17\% of the noun database size).

\begin{figure}[!h]
	\begin{center}
		\includegraphics[scale=0.55]{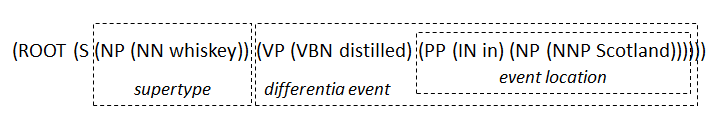} 
		\caption{Syntactic parse tree for the definition of the concept ``Scotch'' and assigned semantic role labels. After being classified as a differentia event, the VP is further analyzed and a PP containing an event location is identified and assigned its own role label.}
		\label{fig:parse}
	\end{center}
\end{figure}

\textbf{Automatic pre-annotation:} the set of 2,000 definitions was automatically pre-annotated according to a rule-based heuristic that takes into account the syntactic patterns identified by statistical analysis as described by Silva et al. \shortcite{silva2016categorization}. Using the Stanford parser \cite{manning2014stanford}, we generated the syntactic parse tree for each definition, identified the relevant phrasal nodes and then assigned the semantic roles more often associated to them. For example: the \textit{supertype} for a noun definition is usually the innermost and leftmost noun phrase (NP) that contains at least one noun (NN); a \textit{differentia event} is usually either a subordinate clause (SBAR) or a verb phrase (VP); an \textit{event location} is normally a prepositional phrase (PP) inside a SBAR or VP and possibly containing a location named entity, and so on. Figure \ref{fig:parse} shows the parse tree generated for the definition of the term ``Scotch'' -- \textit{whiskey distilled in Scotland} -- and the semantic roles automatically assigned to each phrasal node.

\textbf{Data curation:} after the automatic pre-annotation, the definitions were manually curated with the aid of the Brat\footnote{http://brat.nlplab.org/} annotation tool. Misclassifications were fixed and segments missing a role were assigned the appropriate one. Misclassifications and missing roles are due to parser errors or insufficient information (for instance, a PP inside a VP may not contain any named entity, making it hard to correctly distinguish between an event time and an event location). The manual data curation ensured that every segment in each definition, apart of leading determiners and conjunctions between roles (as opposed to conjunctions inside roles), was associated with a semantic role label.

\textbf{Classifier training:} the curated data was then used to train a Recurrent Neural Network (RNN) machine learning model designed for sequence labeling. We used the RNN implementation provided by Mesnil et al. \shortcite{mesnil2015using}, which reports state-of-the-art results for the slot filling task. The dataset was split into training (68\%), validation (17\%) and test (15\%) sets. The best accuracy reached during training was of 80.35\%.

\textbf{Database classification:} the trained classifier was then used to label all WordNet's noun and verb definitions. For simplicity, example sentences and parentheses were excluded from the original glosses. The classification was performed over WordNet 3.0; 82,112 noun definitions and 13,761 verb definitions were labeled.

\begin{figure}[!h]
	\begin{center}
		\includegraphics[scale=0.55]{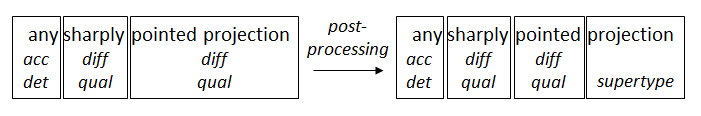} 
		\caption{Classified definition missing a supertype fixed in the post-processing phase.}
		\label{fig:pospproc}
	\end{center}
\end{figure}

\textbf{Data post-processing:} since some of the classified definitions lacked the supertype role, the labeled data had to pass through a post-processing phase. The supertype is a mandatory component in a well-formed definition and, as will be detailed later, the RDF model is structured around it. Following the same syntactic rules adopted for pre-annotation, missing supertypes were identified and the roles around it had its limits adjusted, while the remaining classification was kept unchanged. Figure \ref{fig:pospproc} shows an example of definition (for the term ``spur'') fixed in the post-processing phase.

\begin{figure}[!h]
	\begin{center}
		\includegraphics[width=3.2in, height=1.6in]{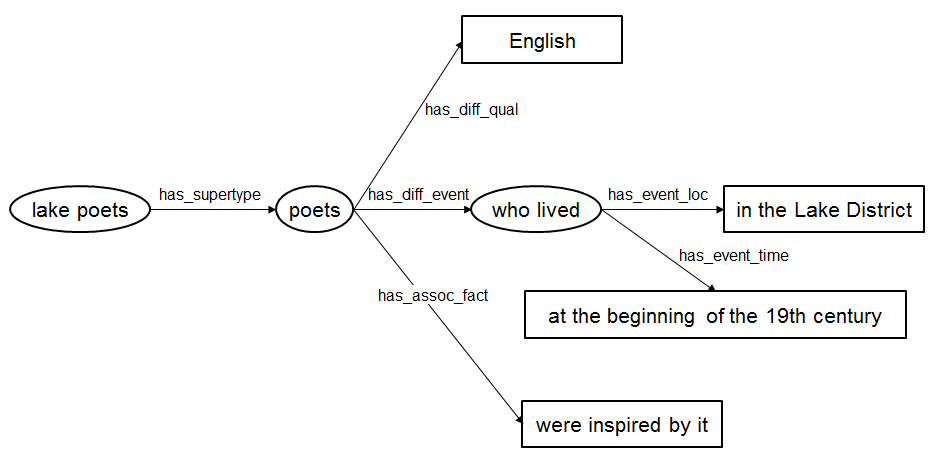}
		\caption{RDF representation for the definition of the ``lake poets'' synset.}
		\label{fig:RDF_example}
	\end{center}	
\end{figure}

\textbf{RDF conversion:} finally, the labeled definitions were serialized in RDF format. In the final graph, a synset is a node and each role in its definition is another node. The synset node is linked to the supertype role, which is, in turn, linked to all the other roles. More specifically, a supertype linked to a role is a \textit{reified} node, and this reified node is linked to the synset node. Reification is also used when a role has components, such as event time and/or location for a differentia event and quality modifier for a differentia quality. In this case, the component is linked to its main role, composing a reified node which is linked to the supertype, creating another reified node which is eventually linked to the synset node. This structure allows the relationships to be fully contextualized. As an example, consider the definition depicted in Figure \ref{fig:classif_example}. The node defined by the concept ``poet'' may be linked to several other nodes in the graph, but it is linked to the differentia quality node ``English'' only in the context of this definition. Supertype nodes are always represented as resources. The differentia quality and differentia event nodes can be represented as either resources, when they have components (event times and/or locations, or quality modifiers) to be linked to, or literals otherwise. All the other roles are represented as literals, and properties are named after role names\footnote{Complete list of the model's properties and namespaces at https://github.com/Lambda-3/WordnetGraph}. Figure \ref{fig:RDF_example} shows the simplified (without reification) RDF representation for the definition in Figure \ref{fig:classif_example}.

Besides WordNetGraph, which is available in both XML and N-Triples format, we provide a set of tools\footnote{https://github.com/ssvivian/DefRelExtractor} that implement the methodology described above. Routines for pre-processing definitions to generate sample data for manual curation, post-processing data returned by a machine learning classifier, and generating the RDF model from the classified data are freely available, along with some auxiliary routines to prepare the data for external tools, such as converting to the \textit{standoff} file format required by the Brat annotation tool and generating a python script that will create the dataset for the RNN classifier.

\section{Application}

WordNetGraph is one of the main components in a text entailment recognition approach aimed at justifying entailment decisions where reasoning over world knowledge is required. Text entailment is defined as a directional relationship between an entailing text \textit{T} and a entailed hypothesis \textit{H}, holding true whenever a human reading T would infer that H is most likely true \cite{dagan2006pascal}. Using WordNetGraph as the world knowledge base, we implemented a navigation algorithm based on distributional semantics \cite{freitas2014distributional} to find  a path in this graph linking T to H, and used the contents of the nodes in this path to build a human-readable justification for the entailment decision. The entailment is rejected if no path is found.

Consider, as an example the entailment pair 39.3 from the BPI dataset\footnote{http://www.cs.utexas.edu/users/$\sim$pclark/bpi-test-suite/}: \\

\noindent 39.3 T: Many cellphones have built-in digital cameras. \\
39.3 H: Many cellphones can take pictures. \\

First, we look for pairs of terms that have a strong semantic relationship and that can prove this entailment to be true, and then send these pairs as input for the graph navigation algorithm. In this example, the best pair is composed by the terms ``digital camera'', which is our \textit{source}, and ``pictures'', our \textit{target}. Starting from the source, we retrieve all the nodes in WordNetGraph linked to it, compute the semantic similarity between each node and the target and choose the one having the highest value as the next node to be visited, and do this recursively until we reach the target. The following segments (triples) are found by the navigation algorithm: 

\begin{center}
$<$digital camera \textit{has\_supertype} camera$>$ \\
$<$camera \textit{has\_supertype} equipment$>$ \\
$<$equipment \textit{has\_diff\_qual} for taking photographs$>$ \\
\end{center}

Since ``photograph'' and ``picture'' are in the same synset node, the search stops at this point, confirming the entailment and producing the following justification, built from the path segments: \\

A digital camera is a kind of camera \\
A camera is an equipment for taking photographs \\
Photograph is synonym of picture \\

Experiments with the BPI dataset and a sample of the Guardian Headlines dataset\footnote{https://goo.gl/4iHdbX} show the results are comparable to those of well-established text entailment algorithms, such as tree edit-distance based \cite{kouylekov2005recognizing} and classification based \cite{wang2008divide}, while providing clear human-like explanations, an important feature still missing in most text entailment recognition approaches. A detailed description of the entailment recognition application, including experiment results and further justification examples can be found in \cite{silva2018recognizing}.

\section{Conclusion}

We presented a method for automatically building a graph world knowledge base from natural language dictionary definitions. Adopting a conceptual model based on entity-centered semantic roles, we trained a supervised machine learning classifier for automatic role labeling and then converted the labeled data into an RDF graph representation. Following this methodology, we created the WordNetGraph, a graph built from the definitions of nouns and verbs in WordNet. A set of tools implementing the methodology is also freely available.

WordNetGraph was successfully used in a text entailment recognition approach based on distributional navigation over definition graphs. Besides using paths in this graph to recognize the entailment, this approach also provides a human-readable justification for the entailment decision. Since each graph node encloses a self-contained amount of information rather than always representing single words, an intelligible justification can be built from a path made up by only a few nodes. As future work, we intend to apply this methodology to other language resources, such as Wiktionary.

\section{Acknowledgments}
Vivian S. Silva is a CNPq Fellow -- Brazil.

\section{Bibliographical References}

\bibliographystyle{lrec}
\bibliography{lrec2018}

\end{document}